%% file: main.tex
\documentclass[sigconf]{acmart}
\AtBeginDocument{%
  \providecommand\BibTeX{{%
    \normalfont B\kern-0.5em{\scshape i\kern-0.25em b}\kern-0.8em\TeX}}}

\setcopyright{acmcopyright}
\copyrightyear{2018}
\acmYear{2018}
\acmDOI{XXXXXXX.XXXXXXX}

\acmPrice{15.00}
\acmISBN{978-1-4503-XXXX-X/18/06}

\usepackage{amsfonts}
\usepackage{algorithm}
\usepackage{algpseudocode}
\usepackage{enumitem}
\usepackage{colortbl}
\usepackage[toc,page]{appendix}
\usepackage{booktabs}
\usepackage{makecell}
\usepackage{times}
\usepackage{soul}
\usepackage{url}
\usepackage[utf8]{inputenc}
\usepackage{graphicx}

\usepackage{amsmath,amssymb,amsthm}
\newtheoremstyle{break}%
  {}{}%
  {\itshape}{}%
  {\bfseries}{}
  {\newline}
  {}%
\theoremstyle{break}
\usepackage{xcolor}
\usepackage{geometry}
\usepackage{hyperref}
\hypersetup{
    colorlinks = false,
    pdfborder = {0 0 0},
}
\usepackage{tikz}
\newcommand{\boxedref}[1]{%
    \tikz[baseline=(char.base)]{%
        \node[shape=rectangle,draw=red,inner sep=2pt] (char) {\hyperref[#1]{\ref*{#1}}};%
    }%
}

\begin{document}

\title[Algorithm for Adaptive Experiments that Trade-off Statistical
Analysis with Reward]{Algorithm for Adaptive Experiments that Trade-off Statistical
Analysis with Reward: Adaptively Combining Uniform Random
Assignment and Thompson Sampling}

\author{Tong Li}
\affiliation{
  \institution{University of Toronto}
  \country{Canada}}
\email{tongli@cs.toronto.edu}

\author{Jacob Nogas}
\affiliation{
  \institution{University of Toronto}
  \country{Canada}
\email{jacob.nogas@mail.utoronto.ca}}

\author{Haochen Song}
\affiliation{
  \institution{University of Toronto}
   \country{Canada}
\email{fred.song@mail.utoronto.ca}}

\author{Anna Rafferty}
\affiliation{
  \institution{Carleton College}
  \country{United States}
\email{arafferty@carleton.edu}}

\author{Eric M. Schwartz}
\affiliation{
  \institution{University of Michigan}
  \country{United States}
\email{ericmsch@umich.edu}}

\author{Audrey Durand}
\affiliation{
  \institution{Université Laval}
  \country{Canada}
\email{audrey.durand@mcgill.ca}}

\author{Harsh Kumar}
\affiliation{
  \institution{University of Toronto}
  \country{Canada}
\email{harsh@cs.toronto.edu}}

\author{Nina Deliu}
\affiliation{
  \institution{University of Cambridge}
  \country{England}
\email{nina.deliu@mrc-bsu.cam.ac.uk}}

\author{Sofia S.~Villar}
\affiliation{
  \institution{University of Cambridge}
  \country{England}
\email{sofia.villar@mrc-bsu.cam.ac.uk}}

\author{Dehan Kong}
\affiliation{
  \institution{University of Toronto}
  \country{Canada}
\email{dehan.kong@utoronto.ca}}

\author{Joseph J.~Williams}
\affiliation{
  \institution{University of Toronto}
  \country{Canada}
\email{williams@cs.toronto.edu}}

\renewcommand{\shortauthors}{Tong, et al.}

\begin{abstract}
Thompson Sampling (TS) has been widely adopted in randomized A/B experiments as an alternative to the traditional uniform randomization (UR) \cite{scott2015multiarmed}. It improves user benefit in the experiment by adaptively randomizing assignment in proportion to the probability they are beneficial. While TS is interpretable and incorporates the randomization key to statistical inference, it can cause biased estimates and increase false positives and false negatives in detecting differences in arm means. We introduce a more ``Statistically Sensitive'' algorithm, TS-PostDiff (Posterior Probability of Small Difference), that mixes TS with UR by using an additional adaptive step, where UR exploration is enacted according to the posterior probability that the difference in arms is `small'. This allows an experimenter to set experiments based on an interpretable `small-difference' threshold, below which it is not that costly to use a traditional UR experiment to get informative data. We evaluate TS-PostDiff against UR, TS, and two other TS variants designed to improve statistical inference. We consider results for experiments across a range of arm differences and sample sizes that incorporate real-world applications into a simulation environment. Our results provide insight into when and why TS-PostDiff and alternative approaches provide better vs.\ worse tradeoffs between benefiting users (reward) and statistical inference (false positive rate and power). TS-PostDiff's adaptivity helps reduce FPR and increase statistical power when differences are small, with less cost on reward when arm difference is big. The work highlights important considerations for future ``Statistically Sensitive'' algorithm development that balances reward and statistical analysis in adaptive experimentation.

\end{abstract}






\maketitle

\section{Introduction}

Randomized ``A/B'' experiments can compare alternative technology-mediated actions or arms to assess their impact on people's behavior. Every day, tens of thousands of these experiments are conducted to compare, e.g., different versions of emails or websites, different components of online education ~\citep{williams2013randomized,heffernan2014assistments,kizilcec2017diverse,kizilcec2020scaling}, or text message categories in mobile health ~\citep{buchholz2013physical,thorsteinsen2014increasing,figueroa2022}. Uniform randomization (UR) are a standard method to learn about the effectiveness of different alternatives. They are typically conducted by allocating participants to arms uniformly at random. However, if the arms being compared differ in how beneficial they are to participants, one might wish to direct more participants to more effective arms. 

Adaptive strategies such as those for solving multi-armed bandit problems offer one potential way to do so~\citep{lomas2016interface,williams2018enhancing}. These algorithms vary the probability that a participant will be assigned to an arm based on the effectiveness of the arms for previous participants. An important benchmark we considered in this paper is Thompson Sampling~\citep[TS;][]{thompson1933likelihood}. It is a Bayesian multi-armed bandit algorithm that maximizes the cumulative ``Rewards'' or positive outcomes, by assigning arms in proportion to their probability of being optimal. 
In addition to being a popular algorithm due to its practicality and asymptotic optimality~\citep{agrawal2012analysis,agrawal2013further,chapelle2011empirical}, TS is based on an intuitive and attractive framework that can be understood at any point as weighted randomization.

When analyzing data from an experiment using statistical hypothesis testing, we typically aim to determine whether conditions differ in their mean effectiveness. This question is important for driving future work: if conditions do differ in effectiveness, the more effective condition can be deployed in broader interventions, and future work can build on this condition. If the conditions do not differ in effectiveness, future work might use cost or other considerations to choose between them, and the reasons why the conditions were hypothesized to differ may need to be revised. 

However, adaptive experiment brings challenges into statistical inference. When data are collected using traditional experimentation, where participants are assigned to conditions uniformly at random, the properties of the hypothesis tests are well known: we can set a parameter to control the false positive rate / type I error (typically controlled at $.05$), and from our beliefs about the likely arm difference, we can determine how many participants are needed to ensure that power (the complement of type II error) reaches a desired threshold (e.g., $0.8$) \citep{cohen1988statistical}. While multi-armed bandit algorithms do a good job maximizing reward, there are well-documented challenges in drawing inferences from adaptively collected data~\citep{xu2013estimation,bowden2017unbiased,nie2018why,shin2019biased,deshpande2018accurate,zhang2020inference,villar2015multiarmed,rafferty2019statistical}. Data collected from these algorithms typically leads to \textit{increased} false positive rates (FPR), meaning that researchers can mistakenly believe that conditions differ and devote considerable time and resources into building on differences that do not exist. This occurs because the algorithm will tend to sample less from an arm when current data has underestimated its effectiveness. Further, data collected from these algorithms typically leads to \textit{decreased} power, where an arm difference is less likely to be detected than if the same number of people participated in a traditional experiment. This occurs because the algorithm tends not to place sufficient participants in both arms to give a confident estimate of the effectiveness of these arms. Lowered power means that more resources need to be devoted to further research to establish the difference between conditions, and may lead to abandoning an intervention that is actually helpful.

Ideally, data scientists and product designers would not have extreme choices in the dichotomy of: (1) reliable statistical inference (typical of UR traditional experiments); and (2) maximizing reward or participant outcomes (typical of TS or other adaptive algorithms). The aim would be to understand how to balance or prioritize reliable statistical inference against Reward maximization.

Motivated by this, the current paper explores how to adjust potentially the most widely used algorithm for adaptively randomization, TS, to address the above trade-off: (1) statistical inference and scientific knowledge--more specifically, high Power to detect differences when they exist and low FPR, against (2) maximizing reward by giving users the most beneficial arms. We introduce an alternative version of TS, named TS-PostDiff (Posterior Probability of Difference), that leverages a \emph{Statistically Sensitive} viewpoint into the Reward maximization framework. Specifically, TS-PostDiff takes a double-Bayesian approach to allocating arms by mixing TS and UR: a participant is allocated to either TS or UR based on the posterior probability that the
difference between two arms is small or large enough, according to a threshold predetermined by domain scientists. 
Large arm differences will prioritize Reward over statistical inference. 

While there is a plethora of TS variants, to our knowledge, none of them have been specifically introduced to address problems in hypothesis testing or to balance the two competing goals. Multi-armed bandit literature that tackled the problem of inference has focused on the introduction of new testing approaches~\citep{zhang2020inference,hadad2021confidence,deliu2021efficient}. Though they have the potential of correcting estimation bias and hypothesis testing FPR, they do not directly address the low power issue caused by the unbalanced sampling from the algorithm itself. Closer to our problem is the literature on best-arm identification~\citep{audibert2010bestarm}, where a TS variant called Top-Two TS~\citep[TT-TS;][]{russo2016simple} has been proposed to increase the accuracy of identifying the best arm. An alternative popular approach is \textit{probability clipping}, that is, constraining randomization probabilities within a range that excludes the extremes~\citep{yao2021powerconstrained}. 

In this work, we focus on the fundamental and ubiquitous scenario of two-arm experiments with binary Rewards, and conduct various experiments to evaluate the proposed TS-PostDiff against alternative TS-based versions for trading-off Reward and FPR/Power. 
In addition to showing the advantages of the proposed solution, 
we provide insights on how and why this occurs in light of the UR-TS mixing. Considerations on the threshold separating the small vs. non-small differences are also provided.



In terms of the organization of the paper, Section \ref{sec:problem_setup} formally defines our problem setup, Section \ref{sec3: related work} covers related work, Section \ref{sec4: intro to algorithsm} proposed TS-PostDiff algorithm, Section \ref{introduce metrics} proposes empirical evaluations of this more ``Statistically Sensitive'' framework, and Section \ref{sec6: discussion} outlines the conclusion and future research directions.

\section{Problem Setting}\label{sec:problem_setup}
In this paper we want to explore adaptive experiments in which we have multiple goals. 

(1) Maximizing reward, the outcomes of participants in an experiment. We focus on real-world adaptive experiments and evaluate the actual reward achieved over a fixed, finite horizon, in contrast to regret analyses that study asymptotic rates.

Formally, assume there are $K$ available experimental conditions, each associated with an underlying reward distribution. These conditions are referred to as \textit{arms}, denoted by $a \in \mathcal{A} = \{1, \dots, K\}$. In the stochastic setting, the reward distribution of each arm is fixed over time. In this work, we consider a two-armed Bernoulli reward problem, and denote the mean reward of arm $a$ as $p_a$. For example, in an educational setting, we can define the reward as $1$ if a student correctly answers a subsequent question after receiving a particular instructional message, and $0$ otherwise.  Assume the experiment has a total length of $T$ (the total number of participants). At each time step $t = 1, \dots, T$, an MAB algorithm $\pi$ selects an arm $a_t \in \{1, \dots, K\}$ and observes a stochastic reward $r_t \sim \text{Bern}(p_{a_t})$.

On the reward side, we want to maximize the cumulative reward benefit \begin{equation}\label{eq:cum_reward}
\sum_{t=1}^n \mathbb{E}[r_{t}],
\end{equation}
which in reality represents improving the average and cumulative benefit achieved across all participants under a given adaptive assignment algorithm.\\

(2) Statistical inference: collecting data that will allow valid inferences about whether or not differences in arms exist. We will quantify this by measuring the type I error / False Positive Rate (FPR) and Power of a statistical test (a two-population Wald test). 

FPR is defined as the probability that we would reject the null hypothesis ($H_0$) when it is true: FPR$=P($reject $H_0 \mid H_0\text{ is true})$; Power refers to the probability that we successfully reject the null hypothesis ($H_0$) when $H_1$ is true: Power$=P($reject $H_0 \mid H_1\text{ is true})$. In our setting, we consider $H_0: p_1 = p_2$ and $H_1: p_1 \neq p_2$.\\


\par
Our goal is to empirically investigate how different algorithms perform on these two goals (reward and hypothesis testing). Thompson Sampling is an algorithm with good reward performance, yet it can greatly jeopardize inference due to its imbalanced sampling. Thus, we explore hybrid algorithms that combine traditional experiments and uniform random with Thompson Sampling, which can have more balanced performance on the two competing objectives. The ideal algorithm would both maximize reward, as measured by equation~\ref{eq:cum_reward}, and have exploration behavior close to UR when arm differences are small or zero to control FPR and Power.

\section{Related Work}
\label{sec3: related work}
Two general approaches have been taken to address poor inference from adaptively collected data: developing methods to make correct inferences directly from data collected using existing adaptive algorithms, or changing the adaptive algorithm to collect better data. 
Work taking the first approach has primarily focused on characterizing bias from ordinary least squares estimates and reducing that bias through a variety of statistical approaches~\cite{xu2013estimation,bowden2017unbiased,deshpande2018accurate,nie2018why,shin2019biased}; reducing the bias of an estimator is related to the final performance of a hypothesis test, but does not directly address the hypothesis testing procedure (typically evaluated through FPR and Power). We focus on statistical hypothesis testing because that is the dominant paradigm for analyzing data in digital field experiments; future work could explore the impact of adding uniform random allocation on Bayesian analysis. There is reason to believe that the issues in incorrect statistical conclusions extend beyond hypothesis tests, as the issues in bias in estimates of means~\cite{nie2018why} and unequal sample sizes are not restricted to one particular kind of test.

More closely related to this paper is work that considers changing the adaptive algorithm to balance competing goals of minimizing regret and obtaining data that is useful for drawing conclusions. In \textit{active exploration} \cite{carpentier2011ucb}, the objective is to estimate the value of all arms as accurately as possible, and thus ensure low estimation errors. The tradeoff between active exploration and maximizing cumulative reward was first addressed using a heuristic algorithm \citep{liu2014trading}, and later formalized as a multi-objective bandit problem integrating rewards and estimation errors~\citep{erraqabi2017trading} which lead to a rigorous, but deterministic, strategy with theoretical guarantees. The work in this paper is complementary to~\cite{erraqabi2017trading} as it focuses on randomized, rather than deterministic strategies, which are likely to be of interest to researchers who typically conduct experiments.

Several recent papers have directly tackled how to collect data that is amenable to statistical hypothesis testing. The Power-constrained bandits algorithm ~\citep{yao2021powerconstrained} explores making multiple allocation decisions for one participant, with the bandit aiming to guarantee the Power of a later hypothesis test while also minimizing regret. Other work~\citep{williamson2017bayesian} uses a finite horizon MDP approach to minimize regret while constraining the minimum number of participants assigned to each arm; they show this leads to improved Power and lower estimation errors compared to an adaptive algorithm without this constraint. $\epsilon$/TT-TS is a high-performing algorithm that introduces a fixed amount of additional exploration of the second-best arm as modeled by TS \citep{russo2016simple}. In contrast to the fixed amount of Uniform Random introduced by $\epsilon$/TT-TS, we introduce the TS-PostDiff algorithm which adaptively varies the amount of Uniform Random and compare it to the state-of-the-art $\epsilon$/TT-TS.

\section{Towards More Statistically Sensitive Thompson Sampling Algorithms}
\label{sec4: intro to algorithsm}
In the section below, we explain the theories and expected outcomes under traditional Thompson Sampling. Following this, we introduce our innovative algorithm, \textit{Thompson Sampling Posterior Probability of Differnece} (TS-PostDiff), which is predicated on incorporating an interpretable degree of Uniform Random allocation: The fundamental premise of TS-PostDiff is to dynamically adjust the quantum of Uniform Random allocation, thereby aimed at more exploration when the differences in the observed treatment effects fall beneath a user-adjustable threshold, and augmented exploitation through Thompson Sampling when such differences surpass the threshold.


We also contrast TS-Postdiff to two other modifications of TS used in applications. First, using a fixed amount $\epsilon$ of UR traditional experimentation embedded in TS, which is also known as Top-Two Thompson Sampling (TT-TS) in the two-armed case, and performs well on problems that aim to identify the best-arm. Second, TS Probability Clipping (TS-ProbClip), a modification that ensure there is always some randomization for supporting statistical inference, by preventing the allocation probability from getting too extreme and approaching 100\%.

\subsection{TS-PostDiff Algorithm}

PostDiff is an adaptation of the well-known algorithm Thompson Sampling (TS). TS-PostDiff aims to achieve superior statistical inference from data, with enhanced Power for best arm identification, and reduced False Positive Rate (FPR), while preserving the reward efficacy of TS. For TS, the quality of inference is particularly poor (as demonstrated in our simulation studies) when the difference between arms is minimal. This issue diminishes as the difference expands (a scenario often easy to make decision in traditional a/b experiment). Consequently, we crafted TS-PostDiff to ensure that the exploration rate is proportional to the likelihood of a significant difference between the top two arms. With this objective in mind, we introduce the TS-PostDiff algorithm, which functions as follows:

\begin{itemize}
    \item with probability $\phi_t$, choose arm uniformly at random.
    \item with probability $1 - \phi_t$, choose arm according to an adaptive allocation strategy (e.g., TS). 
\end{itemize}
We define $\phi_t$ to be the posterior probability (after $t$ steps) that the difference in expected reward between the two arms is less than a threshold $c \in [0,1]$, which we will refer in the future as ``small-difference'' parameter, set by the experimenter. In other words, let
\begin{align}
    \Delta\mathbb{E}(y_t|\textbf{x}_t, \textbf{p}) = 
    |\mathbb{E}(y_t|x_t = 1, p_1) - \mathbb{E}(y_t|x_t = 2, p_2)| \label{eq:deltaE}
\end{align}
then,
\begin{align}
    \phi_t &= \int_{[0,1]^2} \mathbb{I}\big[\Delta\mathbb{E}(y_t|\textbf{x}_t, \textbf{p}) < c\big] \pi(\textbf{p} |\mathcal{D}_t)\ d \textbf{p} \nonumber\\
    &= \int_{[0,1]^2} \mathbb{I}\big[|p_1 - p_2| < c\big] \pi(\textbf{p} |\mathcal{D}_t)\ d \textbf{p}. \label{eq:phi}
\end{align}

where $\mathbb{I}$ is the indicator function, $\textbf{p} = (p_1, p_2)$ is the parameter vector, $\mathcal{D}_t$ is set of available data up to time $t$, and $\pi(\textbf{p}|\mathcal{D}_t)$ is the posterior distribution of the unknown $p_k$'s parameters given the observed data $\mathcal{D}_t$. Specifically, we choose an arm uniformly at random if $|p_1 - p_2| < c$ (with probability $\phi_t$), and use a MAB adaptive allocation strategy otherwise (with probability $1-\phi_t$). 

We can think of $\delta$ as the difference in reward of arms, below which we are willing to forgo reward in favor of improved FPR and Power.

\begin{figure}[t]
\centering
\includegraphics[width=0.95\linewidth]{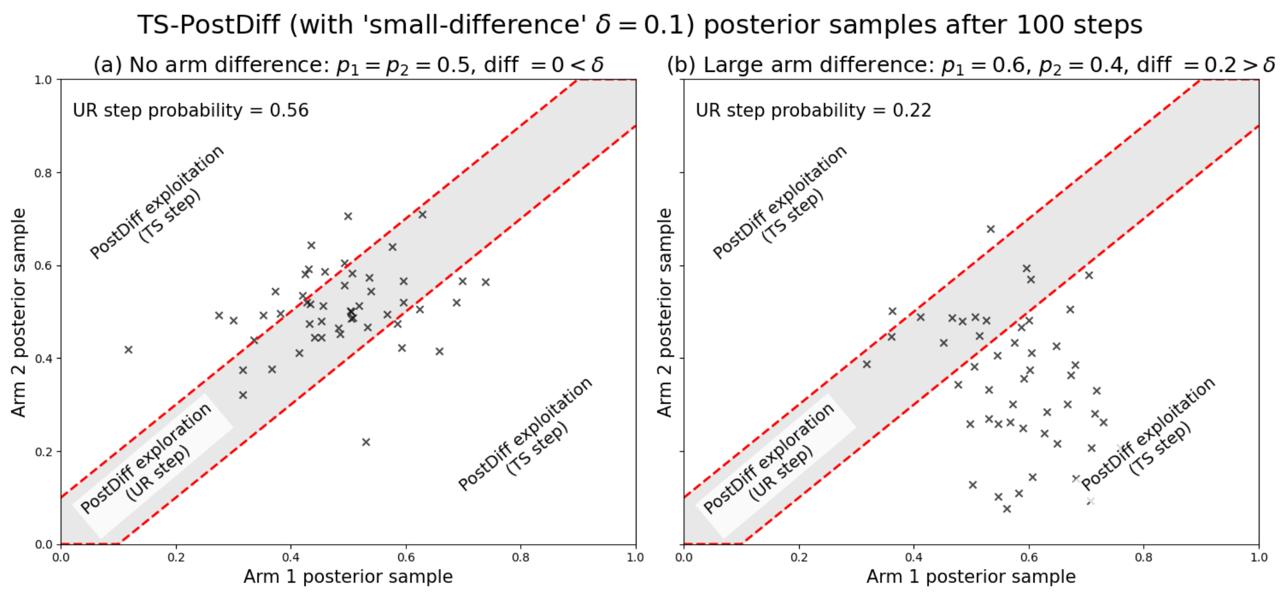}
\caption{
Illustration of TS-PostDiff posterior sampling after $100$ steps in a two-armed Bernoulli bandit.
Each point represents a posterior draw $(\tilde p_1,\tilde p_2)$.
The shaded diagonal band ($|\tilde p_1-\tilde p_2|<\delta$, $\delta=0.1$) indicates the region where TS-PostDiff takes a uniform-randomization (UR) step, while points outside the band correspond to Thompson Sampling (TS).
(\textbf{a}) No arm difference ($p_1=p_2=0.5$), where posterior samples concentrate in the UR region.
(\textbf{b}) Large arm difference ($p_1=0.6$, $p_2=0.4$), where TS-PostDiff predominantly exploits via TS.
}
\label{fig:postdiff_posterior}
\end{figure}

\paragraph{\textbf{Illustration of the posterior difference sampling process}}
We illustrate the modified sampling by visualizing the joint distribution of the p1 and p2, and the decision rule by highlighting which regions correspond to regular TS and which to UR. We show this with two cases of two-armed bandit experiments.

when they are lclose , more sample in middle. (helps wit hexploration)



\begin{algorithm}[tb]
\small
\caption{TS-PostDiff (two-armed case)}
\label{pseudoPSO}
\begin{algorithmic}[1]
\Statex \textbf{Input}: small-difference parameter $\delta$
\Statex \textbf{Initialize}: $\alpha_{k,0}=1,\ \beta_{k,0}=1$ for $k=1,2$
\For {$t=1,2,\ldots $}
    \For {$k = 1,2$}
        \State Sample ${p}_k\sim\text{Beta}(\alpha_k,\beta_k)$
    \EndFor
    \If {$|p_1-p_2|<\delta$}
        \State $x_t = u + 1$, with $u \sim \text{Bern}(1/2)$ 
        \Comment{\emph{\footnotesize{Follow Uniform Random policy}}}
    \Else
        \For {$k = 1,2$}
            \State Sample ${p}_k\sim\text{Beta}(\alpha_k,\beta_k)$ 
            \Comment{\emph{\footnotesize{Resampling step}}}
        \EndFor
        \State $x_t = \arg\max_k {p}_k$  
        \Comment{\emph{\footnotesize{Follow TS policy}}}
    \EndIf
    \State Apply $x_t$ and observe the reward $y_t$
    \State Update posterior for selected arm
\EndFor
\end{algorithmic}
\end{algorithm}

Note the inclusion of a resampling step.\footnote{This step is used to prevent using the same $\textbf{p}$ sample to determine if a difference exceeds the 'small-difference' threshold $\delta$, and for arm selection, as this can result in TS behaving too exploitatively, choosing the estimated best arm too frequently.}

\subsection{Epsilon/Top-Two Thompson Sampling}
$\epsilon$/TT-TS ~\citep{russo2016simple} introduces additional fixed random exploration through a parameter $\beta$, and achieves high performance in best-arm identification. At each time step, $\epsilon$/TT-TS samples from the posterior distribution for each arm, and with probability $\beta$ chooses the arm with the highest sampled value, and with probability $1- \beta$ chooses the arm with second-highest sampled value. With two-arms, this corresponds to allocating with Uniform Random with probability $\epsilon$ and with TS with probability $1-\epsilon$, where $\beta$ = $1 - \frac{\epsilon}{2}$.

\subsection{Thompson Sampling Probability Clipping (TS-Probability Clipping)}
TS-Probability Clipping \citep{zhang2020inference}, adds restriction to the maximum allocation probability to each arm through $prob$\_$max$, which is a parameter takes value between 0.5 and 1. Accordingly, the minimum allocation probability is equal to $1-prob$\_$max$ and this ensures the algorithm always has a chance to explore the seemingly worse arm. In two arm cases, at each time step, TS-Probability Clipping estimate the allocation probability to each arm, and if the maximum allocation probability exceeds $prob$\_$max$, it instead allocates to the better arm with probability $prob$\_$max$ and to the worse arm with probability $1-prob$\_$max$.

\section{Evaluating Algorithms for Adaptive Experiments}

\label{introduce metrics}



Our analysis delves into the possible tradeoff ``range'' between TS and UR with respect to Power and Reward, and we discuss what an optimal or ideal tradeoff might entail. Our findings underscore that TS-PostDiff resonates more closely with the anticipated ideal tradeoff.

As a precursor to understanding the effect of increasing exploration within a reward-maximization algorithm, we evaluate the performance of TT-TS and TS-ProbClip. This examination serves as a foundation for understanding the impact of adding exploration to reward-maximization algorithms.


\subsection{Methods and Definitions}
\label{Sec method and definition}


\paragraph{\textbf{False Positive Rate (FPR) and Statistical Power:}} In our experiments, FPR (i.e., the type I error) and Power (compliment of the type II error) are calculated under the Wald test with a significance level of $0.05$, which is commonly used in web experimentation and in social and behavioral sciences~\citep{fisher1925statistical}. As we discussed earlier, for adaptively collected data, it may cause inflated FPR under regular critical regions.

\paragraph{\textbf{Arm differences (Differences in Arm Means) and Sample Sizes:}} In our simulations, we consider a two-armed Bernoulli bandit setting with arm means denoted as $p_1^{*}$ and $p_2^{*}$. When we say ``arm difference = $w$'', we mean that $p_1^{*} = 0.5 + \frac{w}{2}$, $p_2^{*} = 0.5 - \frac{w}{2}$. We centered $p_1^{*}$ and $p_2^{*}$ at 0.5 to have sufficient space for moving the two arms apart.

In our experiments, we use 197 and 785 as the benchmark sample sizes, as they are the number of samples needed to get a Power of 0.8 using UR algorithm, under arm difference = 0.2 and 0.1, respectively. we choose these as standardized measures of arm differences with appropriate sample size, which were informed by a review of past papers with real-world adaptive experiments. They were also informed by our real-world deployment of TS-PostDiff in an online course. See Appendix \ref{sec C: real-world} for more details.

\paragraph{\textbf{Proportion of Optimal Arm /  'Superior' Arm:}} When the arm difference is greater than zero, the arm with the highest expected reward is designated as the Optimal Arm. Conversely, when the arm difference is zero, the arm with a superior estimated reward mean, attributable to data randomness, is termed the 'Superior Arm', a designation that may vary across different simulation trials. The metric 'Proportion of Optimal Arm / 'Superior' Arm' represents the expected frequency at which an algorithm will select the 'Optimal Arm' or 'Superior Arm' in a given setting. Ideally, an algorithm should allocate more participants to the 'Optimal Arm' to maximize reward, while ensuring an even distribution between the 'Superior' and non-'Superior' Arm to optimize the FPR.

\paragraph{\textbf{Hyper-parameters:}}

Adaptive algorithms result in an inflated false positive rate (FPR), with the extent of this inflation varying according to the actual parameters. To ensure a 'fair' comparison, we focus on algorithm parameters that yield an FPR of either 0.06, 0.07, 0.08, or 0.1, within sample sizes of 197 or 785. Detailed figures related to this are furnished in the Appendix under Section \ref{res}.
 
For all results, we average across the outcomes of $10,000$ simulations.

\subsection{Results}
\label{Section: Result}

Before turning to a more detailed comparison of adaptive algorithms, we begin with a hypothetical empirical example in Section~\ref{sec:postdiff empirical example} to illustrate the practical benefit of PostDiff over classical designs such as UR and TS. This example demonstrates how PostDiff improves the reward--inference trade-off in a concrete educational setting, helping practitioners protect both statistical validity and participant outcomes. \\

After introducing this motivating example, we then provide a more systematic analysis of different adaptive strategies and evaluate their reward and inference tradeoffs across a range of effect sizes. Specifically, we explore three mixture algorithms: (1) TS-PostDiff, our proposed novel algorithm that adapts the amount of exploration based on the probability that the difference between arms falls below a pre-specified gap (or threshold). (2) $\epsilon$/TT-TS, involving a fixed amount of additional UR exploration. (3) TS-ProbClip, employing probability clipping to prevent excessive TS assignment probabilities. 

\subsubsection{Empirical example illustrating the benefit of PostDiff compared to classical design}\label{sec:postdiff empirical example}

\begin{table}[ht]
\centering
\resizebox{1.01\linewidth}{!}{
\begin{tabular}{lccc}
\toprule
 & UR & TS & PostDiff ($c=0.125$) \\
\midrule

\rowcolor{gray!10}
\textbf{No true difference (0.50 vs. 0.50)} \\
Type I error & 0.05 & \textcolor{red}{\textbf{0.13}} & 0.07 \\

\addlinespace[2pt]
\arrayrulecolor{gray!60}\midrule
\arrayrulecolor{black}

\rowcolor{gray!10}
\textbf{Prompt is 10\% more effective} \\
\rowcolor{gray!10}
\textbf{(0.55 vs. 0.45)} \\
Power & 0.803 & \textcolor{red}{\textbf{0.564}} & 0.785 \\
Students assigned to better condition & 393 (50\%) & 678 (86\%) & 541 (69\%) \\
Total students correct & 393 (50\%) & 421 (54\%) & 407 (52\%) \\

\addlinespace[2pt]
\arrayrulecolor{gray!60}\midrule
\arrayrulecolor{black}

\rowcolor{gray!10}
\textbf{Prompt is 20\% more effective} \\
\rowcolor{gray!10}
\textbf{(0.60 vs. 0.40)} \\
Power & 1.000 & 0.991 & 0.997 \\
Students assigned to better condition 
  & \textcolor{red}{\textbf{393 (50\%)}} 
  & 746 (95\%) 
  & 698 (89\%) \\
Total students correct 
  & \textcolor{red}{\textbf{393 (50\%)}} 
  & 463 (59\%) 
  & 455 (58\%) \\

\bottomrule
\end{tabular}
}
\caption{Illustration of how PostDiff can serve as an improved solution to classical designs (TS or UR). In the table we present
results for UR, TS, and PostDiff ($c=0.125$) in the self-explanation prompt experiment with a total sample size of 785 students. 
The three rows correspond to scenarios where the true reward difference between the prompt and no-prompt conditions is 0.00, 0.10, and 0.20. 
We report statistical reliability (Type~I error and Power), the number and percentage of students allocated to the better condition, and the number and percentage who answered correctly. 
Red-bolded entries highlight severe failures of classical designs: inflated Type~I error and low power for TS when the true effect is small, and a large number of sub-optimal assignments under UR when the effect is substantial. 
These results illustrate why more flexible and reliable adaptive designs—such as PostDiff—are needed in practical experimentation settings.
}
\label{tab:selfreflectionresults}
\end{table}

To illustrate how PostDiff improves the reward--inference trade-off in practice, we consider an online-course experiment inspired by the ``Encouraging Self-Reflection'' scenario in \citet{musabirov2025platformadaptive}. The instructor wishes to test whether providing students with a short self-explanation prompt after a multiple-choice question increases the probability of answering the next related question correctly. The experiment compares two conditions: a no-prompt baseline and a self-explanation prompt in which students are asked, ``Can you explain why you chose your answer? Imagine explaining it to another student or your instructor.'' The reward is 1 if the student correctly answers the next question (and 0 otherwise).

PostDiff requires very little statistical background from practitioners. The instructor only needs to specify the threshold $\delta$ that defines what counts as a ``small'' difference between conditions. If the true difference is below this threshold (e.g., $0$ or $0.10$), PostDiff behaves similarly to uniform randomization, keeping students' outcome risk low---something TS fails to do. As shown in Table~\ref{tab:selfreflectionresults}, in these small-effect settings TS produces inflated Type~I error and substantially reduced power, while PostDiff remains close to UR in both statistical reliability and allocation balance.

When the prompt meaningfully improves learning (e.g., a $0.20$ difference), PostDiff automatically shifts toward reward-maximizing behavior. As shown in Table~\ref{tab:selfreflectionresults}, with $785$ students, UR assigns only $393$ students (50\%) to the better condition, whereas PostDiff assigns $698$ students (89\%). This means more than $300$ additional students benefit from the superior intervention, while the method still maintains strong statistical power. In this way, PostDiff offers a simple, practitioner-friendly adaptive design that protects inference when effects are small and protects participants when effects are large.

\subsubsection{\textbf{Evaluating tradeoffs between statistical inference and Reward}}
\begin{table}[tb]
\centering
\caption{
Comparing the power--reward tradeoff in a three-armed Bernoulli experiment.
Each arm’s reward mean $\mu_i$ is sampled independently from $N(0.5, 0.1^2)$ and truncated to $[0,1]$.
We consider a one-sided $t$-test for each arm, testing $H_0: \mu_i = 0.5$ versus $H_1: \mu_i > 0.5$.
All Thompson Sampling (TS) variants achieve higher reward while requiring fewer participants than uniform randomization (UR).
This occurs because adaptive algorithms allocate more samples to better-performing arms, which aligns with the objective of one-sided hypothesis testing.
}
\label{tab:power_reward_tradeoff}
\begin{tabular}{lccc}
\toprule
Algorithm & Parameter & \makecell{Participants to\\ Reach Power} & \makecell{Reward per\\ Participant} \\
\midrule
UR              & --   & 130 & 0.500 \\
TS-PostDiff-N   & 0.08 & 116 & 0.610 \\
$\varepsilon$-TS & 0.37 & 115 & 0.575 \\
TS-ProbClip     & 0.28 & 118 & 0.588 \\
\bottomrule
\end{tabular}
\end{table}

\begin{figure}[ht!]
\centering
\includegraphics[width=85mm]{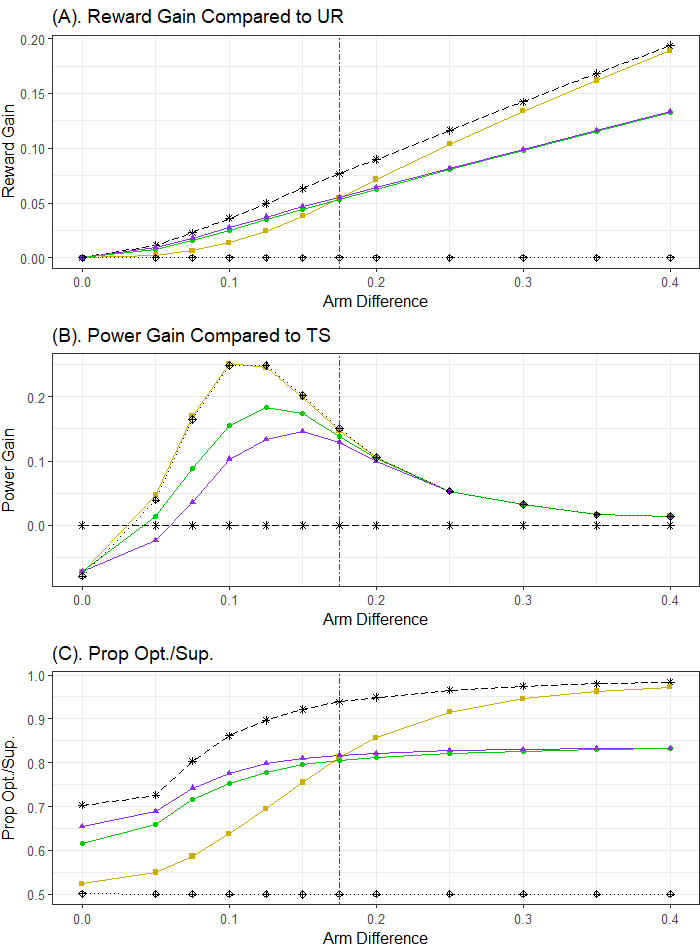}
\Description{Plot showing the evolution of Reward, Power, and Prop Opt./Sup. across different arm differences.}
\includegraphics[width=60mm]{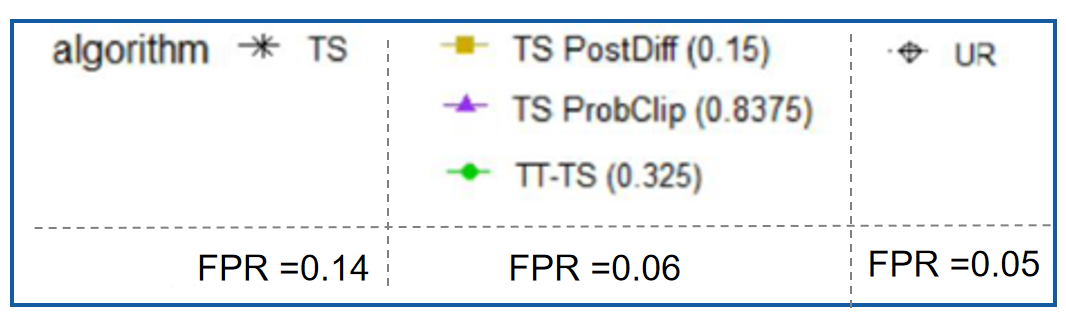}
\caption{Plots showing the evolution of Reward, Power, and Prop Opt./Sup. across differnt arm differences. The sample size is fixed at 785. The algorithm parameters for TS-PostDiff, TT-TS and TS-ProbClip are chosen to align their FPR at 0.06 when there's no arm difference. The red line (at arm difference = 0.175) indicates the location where the Reward from TS-PostDiff catches up with the other two mixture algorithms' Reward performance. In the Power Gain plot, we plot FPR instead when there's no arm difference, which explains why TS comes to the top. }
\label{fig.TS-UR tradeoff}
\end{figure}

In this section, our primary focus is to analyze the tradeoff between Power and Reward across different arm-differences, and discuss what might be an ideal tradeoff.

Figure \ref{fig.TS-UR tradeoff} (Sections (A) and (B)) illustrates how such tradeoff evolves for different algorithm, across a range of arm-differences. For the three mixture algorithms, we chose their parameters to ensure they have the same FPR. It's also important to note that the three mixture algorithms we explore fall between the extremes of Thompson Sampling (TS) and Uniform Random (UR). To make the comparison clearer, we set TS performance as the Power baseline and UR performance as the Reward baseline, as both represent the worst-case scenarios. 

We first examine the performance of Thompson Sampling (TS) and Uniform Randomization (UR), representing the two extreme allocation strategies. When the difference between arms is small, and thus difficult to detect, TS exhibits a substantial loss in power. However, the corresponding reduction in cumulative reward is minimal, as there is relatively little reward advantage to be gained in such settings. As the arm-difference increases, the power advantage of UR diminishes, since the power of all algorithms in this case approaches to 100$\%$. In contrast, the reward advantage of TS grows substantially. This observation motivates the development of an algorithm that behaves more like UR when the arm difference is small, and more like TS when the difference is large.

TS-PostDiff behaves closer to that 'ideal' trend. Comparing to TT-TS or TS-Probclip, when arm-difference is small, we see that TS-PostDiff obtains a Power gain that's much closer to the UR maximum benchmark, with a small sacrifice on reward.  When arm-difference is large, the difference in Power between all those algorithms becomes negligible, while TS-PostDiff performs better in Reward. Moreover, as arm difference increases, the difference in average reward between TS and TS-PostDiff tends to zero, while there's TT-TS or TS-ProbClip fall behind with a linear gap. 

It is worth noting that we can only align FPR for algorithms that have tunable parameters (where the FPR decreases as the parameter increases exploration). This is why TS and UR have FPR levels different from the other algorithms. Because of this, the power curve for UR is slightly under-represented, as it is achieved under a lower FPR. Similarly, the power for TS can be substantially worse (since it is evaluated under a much higher FPR) than what appears in Figure~\ref{fig.TS-UR tradeoff}.

\begin{figure}[ht!]
\centering
\includegraphics[width=85mm]{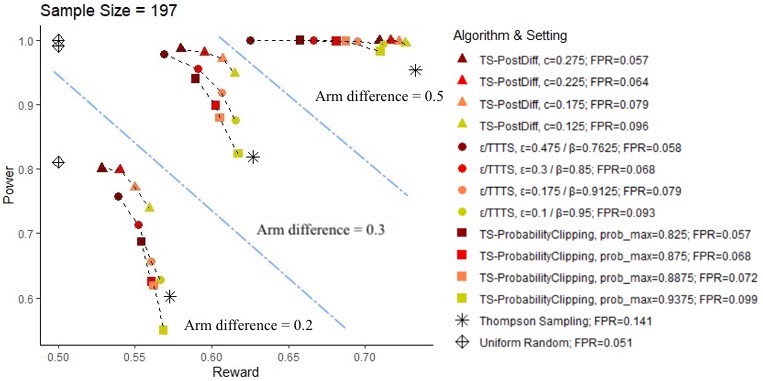}
\includegraphics[width=85mm]{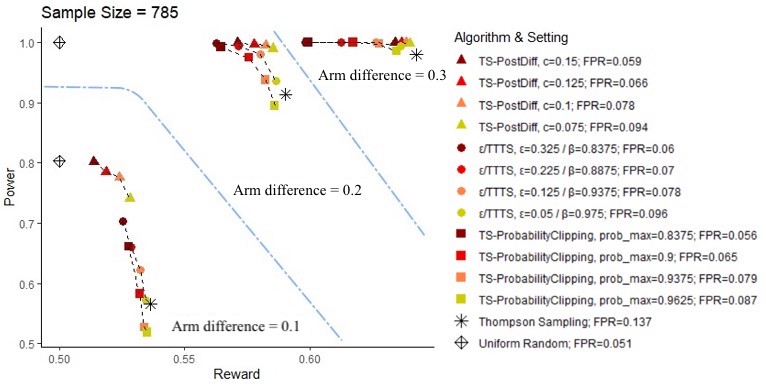}
\caption{Power-reward plots for Uniform Random, TS, TS-PostDiff, $\epsilon$/TT-TS and TS-ProbClip under sample size 197 and 785. The blue dashed lines separate simulation results for the different arm differences/arm differences. \label{figure1}}
\end{figure}

In Figure \ref{figure1}, we present simulation results with a greater generality, which hopefully is a representation of what might happen in real-world deployment. The previous pattern (how the three mixture algorithms compare) exists in those other settings we considered, as well.  

In Table \ref{tab:name3} we average the performance of different algorithms across all of our simulation settings. Reward and Power are averaged from settings where arm difference is greater than zero, while FPR are averaged from settings with no arm difference. Overall, TS-PostDiff has better performance on Reward and Power (FPR is almost the same, by design). This illustrates the advantages we get by using TS-PostDiff, which better aligns with the 'ideal' treadoff pattern we described (doing more exploration when arm difference is small).

\begin{table}
    \centering
    \begin{tabular}{r|r|r|r}
        \toprule
        \multicolumn{1}{c|}{Algorithm}& \multicolumn{1}{c|}{Power}
        & \multicolumn{1}{c|}{Reward}& \multicolumn{1}{c}{FPR}\\
        
        \midrule 
        Uniform Random & 0.930 & 0.500&0.051\\
         
        TS & 0.800 & 0.620&0.139\\
         
        TS-PostDiff & 0.919 & 0.600&0.074\\
         
        $\epsilon$/TT-TS & 0.872 & 0.591&0.075\\
         
        TS-ProbClip & 0.836 & 0.596&0.073\\
        \bottomrule
    \end{tabular}
    \caption{Total average of Power and Mean Reward for Uniform Random, TS, TS-PostDiff, $\epsilon$/TT-TS, and TS-ProbClip across all the different settings of Arm Difference and Sample Size. We also average across different choices of parameters for  TS-PostDiff, $\epsilon$/TT-TS, and TS-ProbClip  such that they produce FPR close to 0.06, 0.07, 0.08 and 0.1 when there's no arm difference ($p_1=p_2=0.5$) under sample size 785 or 195.
    }
    \label{tab:name3}
\end{table}

\subsubsection{\textbf{Algorithm in-depth analysis and comparison}}

To understand the distinction between TS-PostDiff, $\epsilon$/TT-TS, and TS-ProbClip we look at Section (C) in Figure \ref{fig.TS-UR tradeoff}. The definition of 'Prop Sup./Opt.' can be found in Section \ref{Sec method and definition}.

We make analysis from two angles. First, when arm difference is small or zero, TS-PostDiff allocates participants more uniformly to each arm. This is because, in such cases, the posterior difference from the two arms is likely to be small (under the threshold of 'small-difference' parameter '$\delta$'), and hence the UR sampling step from TS-PostDiff mechanism will be triggered more often. Reversely, as arm difference goes large, the Prop Opt. for TS-PostDiff quickly approaches to what TS produces, and that gives TS-PostDiff better reward. For $\epsilon$/TT-TS, and TS-ProbClip, however, the change in terms of Prop Sup./Opt. over difference arm differences is not as significant. In particular, when arm difference is small, they failed to sample the two arms uniformly. For $\epsilon$/TT-TS, this is because the amount of exploration added is upper bounded by $\epsilon$; while for  TS-ProbClip, this is because the 'Probability Clipping' mechanism contributes less when arm difference is small or zero (i.e. it's not quite likely in those cases that the algorithm converges to one arm). On the other hand, when arm difference is large,   $\epsilon$/TT-TS, and TS-ProbClip upper bounds the maximum amount of exploitation of the optimal arm, which prevents them from getting a sub-linear regret.

Now we focus on the single case where arm difference is zero. Here, an interesting fact is: when arm difference is zero, TS-PostDiff samples arms more uniformly (on average), but achieves the same FPR as TT-TS and TS-ProbClip does. This is because: in cases where one arm gets underestimated (due to some bad luck in the data) in the middle of an experiment, the posterior difference between the two arms tends to be big, and hence TS-PostDiff will exploit the 'Superior' arm similar as TS does. In other words, the added UR exploration that TS-PostDiff mechanism brings will decrease by a lot, since the $\phi_t$ in equation \ref{eq:phi} is going to be small. As a result, the poor arm has less chance to recover, and we are more likely to get a falsely significant Wald test score at the end of the experiment. In contrast, the amount of added exploration from TT-TS in this case is not affected. As for TS-PorbClip, it will grants a minimum amount of exploration to the poor arm. Hence, it's arguably 'harder' for TS-PostDiff to get similar FPR with TT-TS and TS-ProbClip. But on the other hand, this 'drawback/flaw' of TS-PostDiff contributes to a better Power for the exact same reason. 

\subsubsection{\textbf{Further analysis on TS-PostDiff regarding the impact of the 'small-arm-difference' parameter '$\delta$'}}
\label{sec: analysis of c}
In the previous section, we compare TS-PostDiff with other algorithms and make analysis when the algorithm parameter '$\delta$' is chosen and fixed. In this section we try to understand the impact of different values of '$\delta$' (the 'small-arm-difference' parameter in TS-PostDiff).

\begin{figure}[!ht]
    \centering
    \includegraphics[width=0.52\textwidth]{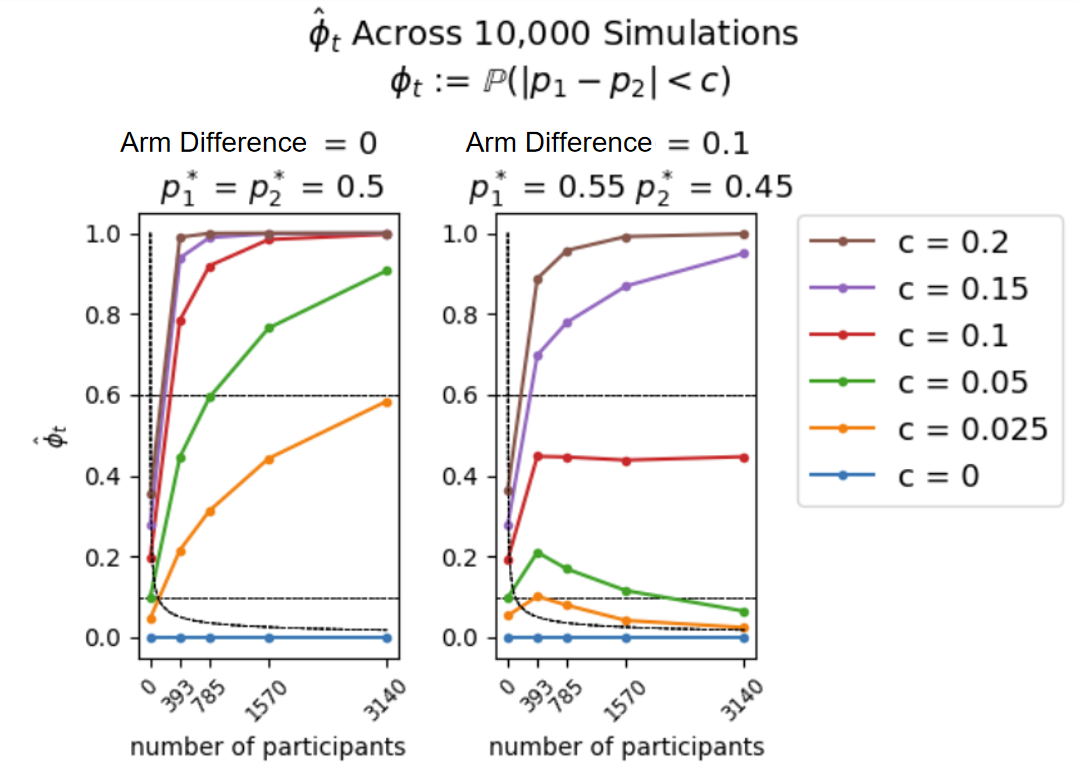}
    \caption{Values of $\hat{\phi_t}$, the estimated probability of choosing actions uniformly, for different sample sizes and values of the 'small-difference' parameter $\delta$ across $10,000$ simulations for the arm difference $0.1$ (right) and $0$ (left). The x-axis denotes sample size and the y-axis shows $\hat{\phi}$. We approximate $\phi$, the true probability of choosing actions uniformly randomly, as $\hat{\phi}$ by taking the proportion of times $|p_1 - p_2| < c$ across $10,000$ simulations, for the given sample size shown on the x-axis. For values of $\delta$ greater than the arm difference, $\hat{\phi}$ converges to $1$, meaning TS-PostDiff allocates actions with Uniform Random. For values of $\delta$ less than the arm difference, $\hat{\phi}$ converges to $0$, meaning TS-PostDiff behaves like standard TS.} 
    
    \label{fig:phi_plots}
\end{figure}

We examine how $\phi_t$ in Equation \ref{eq:phi} evolves as more participants are seen. In Figure \ref{fig:phi_plots}, we show $\hat{\phi_t}$ for various choice of 'small-difference' parameter '$\delta$', for arm differences $0.0$ and $0.1$. 
We approximate $\phi_t$ as $\hat{\phi_t}$ by taking the proportion of times $|p_1 - p_2| < c$ across $10,000$ simulations for the given sample size.
When the 'small-difference' parameter '$\delta$' is above the true arm difference, we see that $\hat{\phi}$ is increasing with sample size towards $1$. When 'small-difference' '$\delta$' is set less than the true arm difference, we see that $\hat{\phi}$ is decreasing towards $0$. These are non-trivial results, as they indicate that the additional UR allocation of TS-PostDiff is able to overcome the bias induced by TS. For example, the above shows that the tendency of TS to lead to overestimating the size of the difference in arms does not prevent $\phi$ from converging to $1$ when the arm difference is $0$.

The above analysis also gives us insight on how to pick the 'small-difference' parameter '$\delta$' when using the TS-PostDiff algorithm. We present detailed discussion in Appendix \ref{sec:guidelines} with examples.

\section{Discussion, Limitations \& Future Work} 
\label{sec6: discussion}
\subsection{Summary}

In digital field experiments, experimenters are often simultaneously concerned with achieving good False Positive Rate/Power and Reward maximization. The issue of balancing reward maximization against statistical inference is important and poses many exciting challenges.

In this work, to better tackle the above-stated problem, we proposed a novel algorithm, TS-PostDiff, that set the parameters considering what 
that one may be less concerned in the experiment stage. It allows the addition of exploration steps to be adaptive to the data, just as TS itself is adaptive to the data (but in a way that has been shown to be potentially problematic). 

TS-PostDiff achieves better overall tradeoff between Reward and statistical inference (Power and FPR) over a range of realistic arm differences, compared to the benchmark algorithms such as TT-TS and TS-ProbClip, and also attains a better theoretical guarantees while maintaining a good simplicity and interoperability. 
Moreover, our evaluation 
can be useful to those interested in adaptive experimentation on the web in a more general way, by offering insights into the strengths and weaknesses of different approaches for adding additional exploration to reward-maximization bandit algorithms, so that practitioners can consider these in choosing which to apply and when.

\subsection{Limitations \& Future Work}

There are limitations to the current paper which point towards future directions. One limitation of the current work is that we focus on two-armed trials. While these are some of the most ubiquitous trial designs, and are highly important in their own right, it could be valuable for future work to understand how an algorithm like TS-PostDiff can be extended to trials with three or more arms.  The complexity can arise in defining hypothesis tests and establishing criteria for "small" arm differences, when the differences between three or more arms might be considered. In addition, we need to figure out: how will combining UR and TS impact the False Positive Rate (FPR), Power, and Reward trade-off as the number of arms increases? For example, when might improving Power to identify the best arm of three or more arms also improve Reward, vs reduce Reward? It's also a harder task in this case to construct simulations that represents a good range of real-world deployments.

Since TS and UR are two widely used approaches, for reward maximization, and statistical analysis, interpolating between them is a natural solution to balancing our competing objectives. But formalizing an objective function for adaptive experimentation algorithms could be extremely beneficial to the development of TS-PostDiff and other algorithms for adaptive experimentation. Defining a function which is a weighted combination of statistical Power, FPR, and expected Reward \cite{erraqabi2017trading} could both provide a common standard by which to compare the performance of different adaptive experimentation algorithms and help experimenters choose an algorithm and set of hyper-parameters based on how important each of Reward, Power, and FPR is to them. Of course, formalizing such an objective function is challenging due to Power requiring assumptions about the likely arm differences, and the wide variety of possible tests, estimators, and experimental goals.

\bibliographystyle{ACM-Reference-Format}
\bibliography{reference}

\input{appendix/appendix}

\end{document}

%% file: appendix/appendix.tex

\newpage
\begin{appendices}

\section{Considerations in choice of $\delta$ for TS-PostDiff}
\label{sec:guidelines}

We now address the question of how to choose the $\delta$ parameter for the TS-PostDiff algorithm. We present two approaches to choosing $\delta$ which reflect different user perspectives.

\begin{figure*}[h!]
    \centering
    \includegraphics[width=\textwidth]{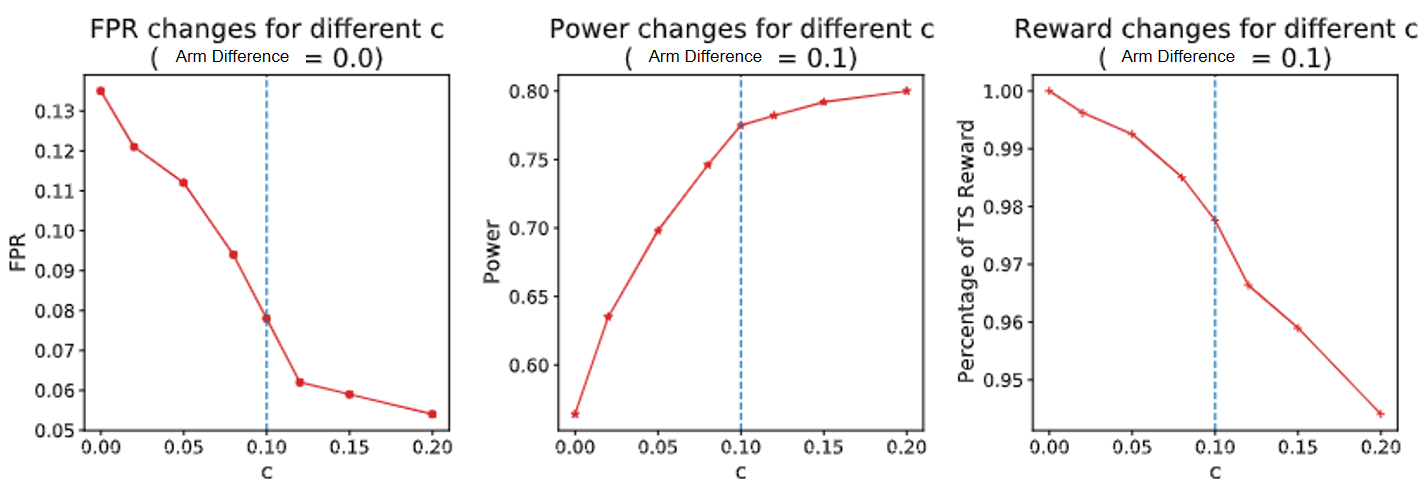}
    \caption{As $\delta$ increases, FPR (left) decreases and Power (middle) increases, while Reward (right)  decreases. Improvements to Power and FPR diminish as $\delta$ increases, while the impact on Reward is roughly linear in $\delta$. Figures show results for a sample size of $785$ simulated participants.  }
    \label{fig:metrics_diff_c}
\end{figure*}

\paragraph{First Recommendation for Choosing $\delta$:}
This discussion of the behaviour of TS-PostDiff in Section \ref{sec: analysis of c} leads us to our first recommendation for choosing $\delta$; we recommend choosing $\delta$ as an arm difference which is small enough that we are willing to forgo reward in favor of improving data analysis capabilities. 

If one is willing to accept $x$ loss in expected reward ($|p_1^* - p_2^*|$) for a sub optimal allocation in favor of improved data analysis, we have strong empirical evidence from Figure \ref{fig:phi_plots} that if we choose $c = x$, when the true arm difference is such that $c > |p_1^* - p_2^*|$, we will converge to always using UR allocation, and if the true arm difference is such that $|p_1^*-p_2^*| < c$, we will converge to TS. In other words, if the true arm difference is below what we are willing accept in expected loss in reward for a sub optimal allocation in favour of improved FPR and Power, TS-PostDiff will likely converge to choosing arms with UR allocation. Similarly, when the true arm difference is greater than what we are willing accept in expected loss in reward for a sub optimal allocation, TS-PostDiff will likely converge to TS.

\paragraph{Example Choosing $\delta$ Based on Admissible Loss in Reward:}
For instance, perhaps we are interested in testing whether design A or design B of a website button results in a greater click through rate (CTR). We will run an experiment with $1000$ users, and we do not want to miss out on an improvement in CTR of over $10\%$ from the optimal design (i.e. design A's CTR is $20\%$ compared to design B's CTR of $10\%$) over the course of our adaptive A/B test, if this led us to reliably discover this arm difference. We would favor discovering the arm difference, since in the short term the decrease in CTR is not very large, but by more reliably discovering relatively small, we can deploy the superior design in the long run where this design choice is used for months, and used by millions of users. We thus set $c = 0.1$ and run the experiment. 

It turns out that the true increase in CTR for the optimal design is $9.6$ \%. Since the arm difference is smaller than what we have deemed admissible through our choice of $\delta$, TS PostDiff will adapt and perform a large amount of UR allocation, thereby increasing our chance of discovering this arm difference, and only facing a decrease in CTR which we have deemed admissible.

By contrast, let's imagine that the difference in click through rate was actually $20\%$, which is greater than what we are willing to give up during the course of our experiment. Since the true arm difference exceeds $\delta$, TS-PostDiff will perform more allocations according to TS than UR allocation, thereby respecting that we would like to improve CTR during the course of our adaptive experiment.


\paragraph{Second Recommendation for Choosing $\delta$:}
 What if one isn’t sure what reward they are willing to give up for improved data analysis? Or, it might be hard for a social-behavioral scientist to make that decision without greater clarity about how much is lost or gained on each dimension of Reward, FPR, and Power. Another approach could be to choose $\delta$ as the best non-zero guess for the true arm difference. 
 
 We motivate the recommendation of choosing $\delta$ as a guess for the true arm difference based on results shown in Figure \ref{fig:metrics_diff_c}, which compares FPR, Power and percentage of TS Reward for a range of $\delta$ values and arm differences of $0$ and $0.1$.
Figure~\ref{fig:metrics_diff_c} shows diminishing returns in Power when $\delta$ exceeds the arm difference of $0.1$, whereas the percentage of TS Reward decreases roughly linearly. FPR also reaches diminishing returns, but a $\delta$ value of $0.1$ is close to the value where we see such diminishing returns. Choosing a $\delta$ value of $0.1$ is thus a reasonable choice, or in general a $\delta$ value equal to the arm difference when arm differences are smaller. If the arm difference is larger though, then the loss in Reward will be greater. Though the percentage of TS Reward is decreasing linearly, if TS Reward is large, then this linear decrease will be more costly. Following the thresholds in behavioural science for small and medium arm differences \cite{cohen1988statistical}, we don't recommend setting $\delta$ higher than $0.2$.

In sum, we can view the choice of $\delta$ from two perspectives. We can use $\delta$ to minimize using TS when arm differences are small enough that we would favour improved inference over reward maximization, and we can also use $\delta$ as a way to influence how TS-PostDiff trades-off between reward maximization and Uniform Random exploration. These two perspectives (and a combination of them) can thus guide a user in choosing $\delta$.

\section{Real-World Adaptive Experiments}
\label{sec C: real-world}
\begin{figure}[ht!]
\includegraphics[width=80mm]{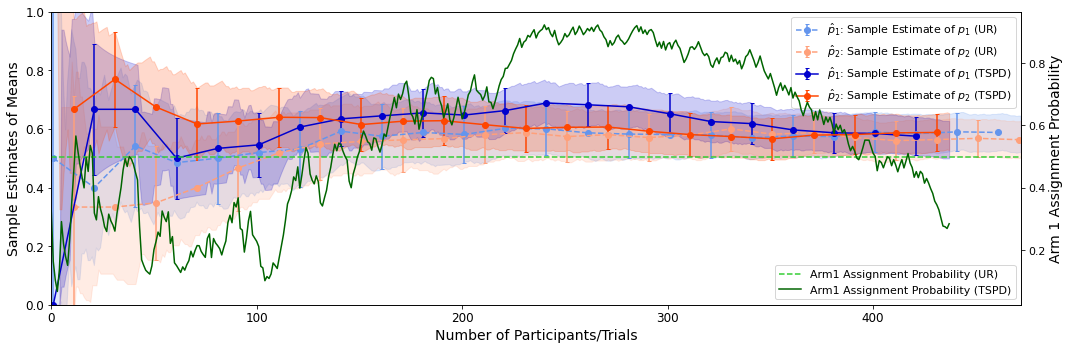}
\includegraphics[width=80mm]{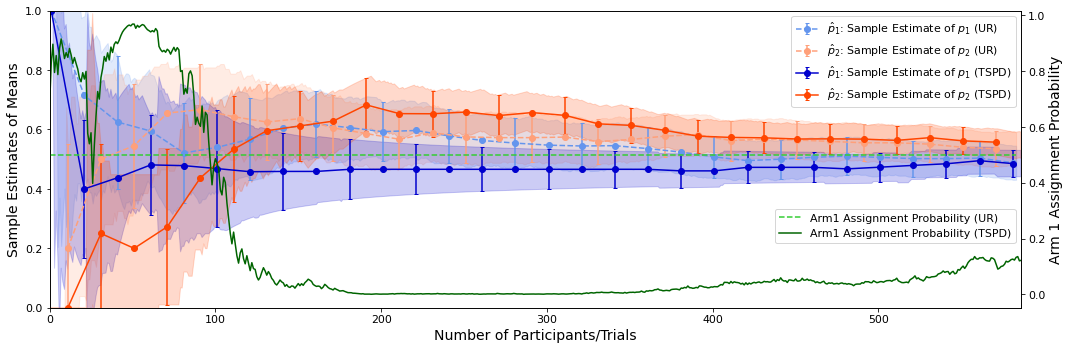}
\caption{Real world deployments where we tested the behavior of TS-PostDiff while comparing it with Uniform Random allocation. \textit{Case 1 (Top)}: 473 participants were randomly allocated to the arms, showing an arm difference of $0.016$; the other 438 students were allocated by a TS-PostDiff policy with $c = 0.1$, getting an arm difference of $0.27$, showing that TS-PostDiff chose to explore given that the estimated arm difference was small compared with the parameter $\delta$. \textit{Case 2 (Bottom)}: 584 participants were randomly allocated to the arms, showing an arm difference of $0.045$; the other 587 students were allocated by a TS-PostDiff policy with $c = 0.02$, getting an arm difference of $0.062$, showing that TS-PostDiff chose to exploit the arm with better mean given that the estimated arm difference was large compared with the parameter $\delta$. \label{fig:real_world_data}}
\end{figure}

We considered a range of real-world online experiments that could be adapted using algorithms that try to trade-off helping participants with statistical analysis \cite{bakshy2014deploying}. These ranged from website conversions \cite{wang2022adaptive}, messages to get people vaccinated \cite{milkman2022680} \cite{kumar2022sounds}, and optimizing instructor explanations \cite{williams2018enhancing}. We conducted 10 real-world educational experiments during 2021, where we collected data side by side using both UR (traditional RCT/experiment) and an adaptive experiment using the TS-PostDiff algorithm. Figure~\ref{fig:real_world_data} shows illustrative examples from these real-world deployments. To evaluate the performance of different algorithms for statistical analysis, we need to know how often these cases occur: specifically, the false positive rate and the power. Case (1) is TS-PostDiff producing closer to Uniform Random allocation when there is little evidence for a difference in arm means, reducing the chances of getting a false positive with no—or minimal—cost to reward; (2) TS-PostDiff doing TS reward maximization when there seems to be a difference in arm means that is \textit{large} relative to the parameter for \textit{small} differences (below which maximizing reward is not a high priority). 

Of course, a few real-world examples of case studies does not reveal the behaviour of the algorithms over many adaptive experiments. We therefore specified ground truth scenarios with arm differences and sample sizes, using these to investigate through simulation each algorithm's behaviour over 10,000 experiments.

\begin{table}
    \centering
    \begin{tabular}{r|r|r}
        \toprule
        \multicolumn{3}{c}{10,000 simulations}\\
        \multicolumn{3}{c}{n = 1,171}\\
        \multicolumn{3}{c}{FPR = 0.06}\\
        
        \midrule 
        
        \multicolumn{1}{c|}{Algorithm}
        & \multicolumn{1}{c|}{Reward}
        & \multicolumn{1}{c}{Power}\\
        
        \midrule 
        TS-PostDiff ($\delta$ = 0.12) & 0.523(0.000) & 0.458(0.005)\\
         
        $\epsilon$/TT-TS ($\epsilon$ = 0.34) & 0.525(0.000) & 0.392(0.005)\\
         
        TS-ProbClip ($p\_max$ = 0.83) & 0.524(0.000) & 0.365(0.005)\\
        \bottomrule
    \end{tabular}
    \caption{Power and Reward per algorithm calculated from $10,000$ simulations using rewards from a real-world deployment. The real-world data consists of $1,171$ binary rewards from an educational setting that showed an effect size of $0.045$. We calculated the specific parameter that would yield an FPR of approximately 6\%. We show that in this scenario with small effect size, by trading off 0.38\% in Reward compared with $\epsilon$/TT-TS, TS-PostDiff increased Power by 16.84\%, or 0.19\% in Reward for 25.48\% compared with TS-ProbClip.
    }
    \label{tab:real_world_simulation}
\end{table}

To preview, Table \ref{tab:real_world_simulation} shows suggestive evidence of TS-PostDiff's performance compared with other algorithms (i.e. $\epsilon$/TT-TS and TS-ProbClip), using bootstrap sampling from one of the real-world experiments. When FPR is fixed to 0.06, for a minimal/negligible reduction in Reward, it achieves a considerably better Power, because more UR data collection allows for better detecting such a small effect. We now present full methodology behind these results and the more comprehensive simulations. 

 We then used the effect sizes/arm differences and sample sizes of these real-world experiments to guide our choice of a simulation environment to compare TS-postdiff to $\epsilon$/TT-TS, TS-Probability Clipping, TS, and UR.

\section{Detailed simulation results}
\label{res}

Appendix Table \ref{tab:name1} and Appendix Table \ref{tab:name2} summarizes simulation results used in Figure \ref{figure1}. 

\begin{table*}[!ht]
    \centering
    \begin{tabular}{c|cc|cc|cc|cc|cc}
        \toprule
        \multicolumn{11}{c}{n = 197}\\
        \bottomrule
        
        \toprule
        \multicolumn{11}{c}{arm difference = 0.2}\\
        
        \midrule
        
        \multicolumn{1}{c|}{Algorithm}
        & \multicolumn{5}{c|}{Power}
        & \multicolumn{5}{c}{Reward}\\
        
        \midrule
        
        \multicolumn{1}{c|}{Uniform Random}
        & \multicolumn{5}{c|}{0.804}
        & \multicolumn{5}{c}{0.500}\\
        
        \midrule
        
        \multicolumn{1}{c|}{TS}
        & \multicolumn{5}{c|}{0.586}
        & \multicolumn{5}{c}{0.572}\\
        
        \midrule
        
        \multicolumn{1}{c|}{Algorithm}
        & \multicolumn{2}{c|}{FPR = 0.06}
        & \multicolumn{2}{c|}{FPR = 0.07}
        & \multicolumn{2}{c|}{FPR = 0.08}
        & \multicolumn{2}{c|}{FPR = 0.1}
        & \multicolumn{2}{c}{Mean across FPRs}\\
        
        \cmidrule[\heavyrulewidth]{2-11}
        
        & Power
        & \multicolumn{1}{c|}{Reward}
        & Power
        & \multicolumn{1}{c|}{Reward}
        & Power
        & \multicolumn{1}{c|}{Reward}
        & Power
        & \multicolumn{1}{c|}{Reward}
        & Power                 
        & Reward\\
        
        \midrule
        
        \multicolumn{1}{c|}{TS-PostDiff}
        & 0.801
        & \multicolumn{1}{c|}{0.529}
        & 0.798
        & \multicolumn{1}{c|}{0.540} 
        & 0.772
        & \multicolumn{1}{c|}{0.550} 
        & 0.739
        & \multicolumn{1}{c|}{0.559} 
        & 0.777
        & 0.545 \\
        
        \cmidrule[\heavyrulewidth]{2-9}
        
        \multicolumn{1}{c|}{}
        & \multicolumn{2}{c|}{$\delta$ = 0.275}
        & \multicolumn{2}{c|}{$\delta$ = 0.225}
        & \multicolumn{2}{c|}{$\delta$ = 0.175}
        & \multicolumn{2}{c|}{$\delta$ = 0.125}
        &
        &\\
        
        \midrule
        
        \multicolumn{1}{c|}{}
        & 0.758
        & \multicolumn{1}{c|}{0.539} 
        & 0.713
        & \multicolumn{1}{c|}{0.552} 
        & 0.657
        & \multicolumn{1}{c|}{0.561} 
        & 0.628
        & \multicolumn{1}{c|}{0.566} 
        & 0.689
        & 0.555\\
        
        \cmidrule[\heavyrulewidth]{2-9}
        
        \multicolumn{1}{c|}{$\epsilon$/TT-TS}
        & \multicolumn{2}{c|}{$\beta$ = 0.7625}
        & \multicolumn{2}{c|}{$\beta$ = 0.85}
        & \multicolumn{2}{c|}{$\beta$ = 0.9125}
        & \multicolumn{2}{c|}{$\beta$ = 0.95}
        &
        &\\ 
        
        \multicolumn{1}{c|}{}
        & \multicolumn{2}{c|}{$\epsilon$ = 0.465}
        & \multicolumn{2}{c|}{$\epsilon$ = 0.3}
        & \multicolumn{2}{c|}{$\epsilon$ = 0.175}
        & \multicolumn{2}{c|}{$\epsilon$ = 0.1}
        &
        &\\ 
        
        \midrule
        
        \multicolumn{1}{c|}{TS-ProbClip}
        & 0.687
        & \multicolumn{1}{c|}{0.554} 
        & 0.625
        & \multicolumn{1}{c|}{0.561} 
        & 0.619
        & \multicolumn{1}{c|}{0.562} 
        & 0.549
        & \multicolumn{1}{c|}{0.569} 
        & 0.620
        & 0.562 \\
        
        \cmidrule[\heavyrulewidth]{2-9}
        
        \multicolumn{1}{c|}{}
        & \multicolumn{2}{c|}{prob\_max = 0.825}
        & \multicolumn{2}{c|}{prob\_max = 0.875}
        & \multicolumn{2}{c|}{prob\_max = 0.8875}
        & \multicolumn{2}{c|}{prob\_max = 0.9375}
        &
        &\\
        
        \bottomrule

        \toprule

        \multicolumn{11}{c}{arm difference = 0.3}\\
        
        \midrule
        
        \multicolumn{1}{c|}{Algorithm}
        & \multicolumn{5}{c|}{Power}
        & \multicolumn{5}{c}{Reward}\\
        
        \midrule
        
        \multicolumn{1}{c|}{Uniform Random}
        & \multicolumn{5}{c|}{0.991}
        & \multicolumn{5}{c}{0.500}\\
        
        \midrule
        
        \multicolumn{1}{c|}{TS}
        & \multicolumn{5}{c|}{0.810}
        & \multicolumn{5}{c}{0.627}\\
        
        \midrule
        
        \multicolumn{1}{c|}{Algorithm}
        & \multicolumn{2}{c|}{FPR = 0.06}
        & \multicolumn{2}{c|}{FPR = 0.07}
        & \multicolumn{2}{c|}{FPR = 0.08}
        & \multicolumn{2}{c|}{FPR = 0.1}
        & \multicolumn{2}{c}{Mean across FPRs}\\
        
        \cmidrule[\heavyrulewidth]{2-11}
        
        & Power
        & \multicolumn{1}{c|}{Reward}
        & Power
        & \multicolumn{1}{c|}{Reward}
        & Power
        & \multicolumn{1}{c|}{Reward}
        & Power
        & \multicolumn{1}{c|}{Reward}
        & Power                 
        & Reward\\
        
        \midrule
        
        \multicolumn{1}{c|}{TS-PostDiff}
        & 0.986
        & \multicolumn{1}{c|}{0.580}
        & 0.981
        & \multicolumn{1}{c|}{0.595} 
        & 0.971
        & \multicolumn{1}{c|}{0.607} 
        & 0.948
        & \multicolumn{1}{c|}{0.615} 
        & 0.972
        & 0.599 \\
        
        \cmidrule[\heavyrulewidth]{2-9}
        
        \multicolumn{1}{c|}{}
        & \multicolumn{2}{c|}{$\delta$ = 0.275}
        & \multicolumn{2}{c|}{$\delta$ = 0.225}
        & \multicolumn{2}{c|}{$\delta$ = 0.175}
        & \multicolumn{2}{c|}{$\delta$ = 0.125}
        &
        &\\
        
        \midrule
        
        \multicolumn{1}{c|}{}
        & 0.978
        & \multicolumn{1}{c|}{0.569} 
        & 0.955
        & \multicolumn{1}{c|}{0.591} 
        & 0.918
        & \multicolumn{1}{c|}{0.607} 
        & 0.876
        & \multicolumn{1}{c|}{0.616} 
        & 0.932
        & 0.596\\
        
        \cmidrule[\heavyrulewidth]{2-9}
        
        \multicolumn{1}{c|}{$\epsilon$/TT-TS}
        & \multicolumn{2}{c|}{$\beta$ = 0.7625}
        & \multicolumn{2}{c|}{$\beta$ = 0.85}
        & \multicolumn{2}{c|}{$\beta$ = 0.9125}
        & \multicolumn{2}{c|}{$\beta$ = 0.95}
        &
        &\\ 
        
        \multicolumn{1}{c|}{}
        & \multicolumn{2}{c|}{$\epsilon$ = 0.465}
        & \multicolumn{2}{c|}{$\epsilon$ = 0.3}
        & \multicolumn{2}{c|}{$\epsilon$ = 0.175}
        & \multicolumn{2}{c|}{$\epsilon$ = 0.1}
        &
        &\\ 
        
        \midrule
        
        \multicolumn{1}{c|}{TS-ProbClip}
        & 0.940
        & \multicolumn{1}{c|}{0.590} 
        & 0.898
        & \multicolumn{1}{c|}{0.603} 
        & 0.879
        & \multicolumn{1}{c|}{0.605} 
        & 0.824
        & \multicolumn{1}{c|}{0.617} 
        & 0.885
        & 0.604 \\
        
        \cmidrule[\heavyrulewidth]{2-9}
        
        \multicolumn{1}{c|}{}
        & \multicolumn{2}{c|}{prob\_max = 0.825}
        & \multicolumn{2}{c|}{prob\_max = 0.875}
        & \multicolumn{2}{c|}{prob\_max = 0.8875}
        & \multicolumn{2}{c|}{prob\_max = 0.9375}
        &
        &\\
        
        \bottomrule

        \toprule

        \multicolumn{11}{c}{arm difference = 0.5}\\
        
        \midrule
        
        \multicolumn{1}{c|}{Algorithm}
        & \multicolumn{5}{c|}{Power}
        & \multicolumn{5}{c}{Reward}\\
        
        \midrule
        
        \multicolumn{1}{c|}{Uniform Random}
        & \multicolumn{5}{c|}{1.000}
        & \multicolumn{5}{c}{0.500}\\
        
        \midrule
        
        \multicolumn{1}{c|}{TS}
        & \multicolumn{5}{c|}{0.954}
        & \multicolumn{5}{c}{0.733}\\
        
        \midrule
        
        \multicolumn{1}{c|}{Algorithm}
        & \multicolumn{2}{c|}{FPR = 0.06}
        & \multicolumn{2}{c|}{FPR = 0.07}
        & \multicolumn{2}{c|}{FPR = 0.08}
        & \multicolumn{2}{c|}{FPR = 0.1}
        & \multicolumn{2}{c}{Mean across FPRs}\\
        
        \cmidrule[\heavyrulewidth]{2-11}
        
        & Power
        & \multicolumn{1}{c|}{Reward}
        & Power
        & \multicolumn{1}{c|}{Reward}
        & Power
        & \multicolumn{1}{c|}{Reward}
        & Power
        & \multicolumn{1}{c|}{Reward}
        & Power                 
        & Reward\\
        
        \midrule
        
        \multicolumn{1}{c|}{TS-PostDiff}
        & 1.000
        & \multicolumn{1}{c|}{0.709}
        & 0.999
        & \multicolumn{1}{c|}{0.717} 
        & 0.998
        & \multicolumn{1}{c|}{0.722} 
        & 0.995
        & \multicolumn{1}{c|}{0.726} 
        & 0.998
        & 0.719 \\
        
        \cmidrule[\heavyrulewidth]{2-9}
        
        \multicolumn{1}{c|}{}
        & \multicolumn{2}{c|}{$\delta$ = 0.275}
        & \multicolumn{2}{c|}{$\delta$ = 0.225}
        & \multicolumn{2}{c|}{$\delta$ = 0.175}
        & \multicolumn{2}{c|}{$\delta$ = 0.125}
        &
        &\\
        
        \midrule
        
        \multicolumn{1}{c|}{}
        & 1.000
        & \multicolumn{1}{c|}{0.625} 
        & 1.000
        & \multicolumn{1}{c|}{0.666} 
        & 0.999
        & \multicolumn{1}{c|}{0.695} 
        & 0.994
        & \multicolumn{1}{c|}{0.712} 
        & 0.998
        & 0.675\\
        
        \cmidrule[\heavyrulewidth]{2-9}
        
        \multicolumn{1}{c|}{$\epsilon$/TT-TS}
        & \multicolumn{2}{c|}{$\beta$ = 0.7625}
        & \multicolumn{2}{c|}{$\beta$ = 0.85}
        & \multicolumn{2}{c|}{$\beta$ = 0.9125}
        & \multicolumn{2}{c|}{$\beta$ = 0.95}
        &
        &\\ 
        
        \multicolumn{1}{c|}{}
        & \multicolumn{2}{c|}{$\epsilon$ = 0.465}
        & \multicolumn{2}{c|}{$\epsilon$ = 0.3}
        & \multicolumn{2}{c|}{$\epsilon$ = 0.175}
        & \multicolumn{2}{c|}{$\epsilon$ = 0.1}
        &
        &\\
        
        \midrule
        
        \multicolumn{1}{c|}{TS-ProbClip}
        & 1.000
        & \multicolumn{1}{c|}{0.658} 
        & 0.999
        & \multicolumn{1}{c|}{0.681} 
        & 0.998
        & \multicolumn{1}{c|}{0.687} 
        & 0.983
        & \multicolumn{1}{c|}{0.710} 
        & 0.995
        & 0.684 \\
        
        \cmidrule[\heavyrulewidth]{2-9}
        
        \multicolumn{1}{c|}{}
        & \multicolumn{2}{c|}{prob\_max = 0.825}
        & \multicolumn{2}{c|}{prob\_max = 0.875}
        & \multicolumn{2}{c|}{prob\_max = 0.8875}
        & \multicolumn{2}{c|}{prob\_max = 0.9375}
        &
        &\\
        
        \bottomrule

    \end{tabular}
    \caption{Comparison of Power and Mean Reward for Uniform Random, TS, TS-PostDiff, $\epsilon$/TT-TS, and TS-ProbClip with sample size = 197 and three arm differences (0.2, 0.3, and 0.5). For TS-PostDiff, $\epsilon$/TT-TS, and TS-ProbClip, we choose four different parameters for each of them that gives a false positive rate of 0.06, 0.07, 0.08 and 0.1, respectively. 
    }
    \label{tab:name1}
\end{table*}

\begin{table*}[!ht]
    \centering
    \begin{tabular}{c|cc|cc|cc|cc|cc}
        \toprule
        \multicolumn{11}{c}{n = 785}\\
        \bottomrule
        
        \toprule
        \multicolumn{11}{c}{arm difference = 0.1}\\
        
        \midrule
        
        \multicolumn{1}{c|}{Algorithm}
        & \multicolumn{5}{c|}{Power}
        & \multicolumn{5}{c}{Reward}\\
        
        \midrule
        
        \multicolumn{1}{c|}{Uniform Random}
        & \multicolumn{5}{c|}{0.803}
        & \multicolumn{5}{c}{0.500}\\
        
        \midrule
        
        \multicolumn{1}{c|}{TS}
        & \multicolumn{5}{c|}{0.564}
        & \multicolumn{5}{c}{0.536}\\
        
        \midrule
        
        \multicolumn{1}{c|}{Algorithm}
        & \multicolumn{2}{c|}{FPR = 0.06}
        & \multicolumn{2}{c|}{FPR = 0.07}
        & \multicolumn{2}{c|}{FPR = 0.08}
        & \multicolumn{2}{c|}{FPR = 0.1}
        & \multicolumn{2}{c}{Mean across FPRs}\\
        
        \cmidrule[\heavyrulewidth]{2-11}
        
        & Power
        & \multicolumn{1}{c|}{Reward}
        & Power
        & \multicolumn{1}{c|}{Reward}
        & Power
        & \multicolumn{1}{c|}{Reward}
        & Power
        & \multicolumn{1}{c|}{Reward}
        & Power                 
        & Reward\\
        
        \midrule
        
        \multicolumn{1}{c|}{TS-PostDiff}
        & 0.801
        & \multicolumn{1}{c|}{0.514}
        & 0.785
        & \multicolumn{1}{c|}{0.519} 
        & 0.776
        & \multicolumn{1}{c|}{0.524} 
        & 0.741
        & \multicolumn{1}{c|}{0.528} 
        & 0.776
        & 0.521 \\
        
        \cmidrule[\heavyrulewidth]{2-9}
        
        \multicolumn{1}{c|}{}
        & \multicolumn{2}{c|}{$\delta$ = 0.15}
        & \multicolumn{2}{c|}{$\delta$ = 0.125}
        & \multicolumn{2}{c|}{$\delta$ = 0.1}
        & \multicolumn{2}{c|}{$\delta$ = 0.075}
        &
        &\\
        
        \midrule
        
        \multicolumn{1}{c|}{}
        & 0.704
        & \multicolumn{1}{c|}{0.525} 
        & 0.660
        & \multicolumn{1}{c|}{0.529} 
        & 0.622
        & \multicolumn{1}{c|}{0.532} 
        & 0.572
        & \multicolumn{1}{c|}{0.535} 
        & 0.639
        & 0.530\\
        
        \cmidrule[\heavyrulewidth]{2-9}
        
        \multicolumn{1}{c|}{$\epsilon$/TT-TS}
        & \multicolumn{2}{c|}{$\beta$ = 0.8375}
        & \multicolumn{2}{c|}{$\beta$ = 0.8875}
        & \multicolumn{2}{c|}{$\beta$ = 0.9375}
        & \multicolumn{2}{c|}{$\beta$ = 0.975}
        &
        &\\ 
        
        \multicolumn{1}{c|}{}
        & \multicolumn{2}{c|}{$\epsilon$ = 0.325}
        & \multicolumn{2}{c|}{$\epsilon$ = 0.225}
        & \multicolumn{2}{c|}{$\epsilon$ = 0.125}
        & \multicolumn{2}{c|}{$\epsilon$ = 0.05}
        &
        &\\
        
        \midrule
        
        \multicolumn{1}{c|}{TS-ProbClip}
        & 0.661
        & \multicolumn{1}{c|}{0.528} 
        & 0.583
        & \multicolumn{1}{c|}{0.532} 
        & 0.527
        & \multicolumn{1}{c|}{0.534} 
        & 0.518
        & \multicolumn{1}{c|}{0.535} 
        & 0.572
        & 0.532 \\
        
        \cmidrule[\heavyrulewidth]{2-9}
        
        \multicolumn{1}{c|}{}
        & \multicolumn{2}{c|}{prob\_max = 0.8375}
        & \multicolumn{2}{c|}{prob\_max = 0.9}
        & \multicolumn{2}{c|}{prob\_max = 0.9375}
        & \multicolumn{2}{c|}{prob\_max = 0.9625}
        &
        &\\
        
        \bottomrule

        \toprule

        \multicolumn{11}{c}{arm difference = 0.2}\\
        
        \midrule
        
        \multicolumn{1}{c|}{Algorithm}
        & \multicolumn{5}{c|}{Power}
        & \multicolumn{5}{c}{Reward}\\
        
        \midrule
        
        \multicolumn{1}{c|}{Uniform Random}
        & \multicolumn{5}{c|}{1.000}
        & \multicolumn{5}{c}{0.500}\\
        
        \midrule
        
        \multicolumn{1}{c|}{TS}
        & \multicolumn{5}{c|}{0.911}
        & \multicolumn{5}{c}{0.590}\\
        
        \midrule
        
        \multicolumn{1}{c|}{Algorithm}
        & \multicolumn{2}{c|}{FPR = 0.06}
        & \multicolumn{2}{c|}{FPR = 0.07}
        & \multicolumn{2}{c|}{FPR = 0.08}
        & \multicolumn{2}{c|}{FPR = 0.1}
        & \multicolumn{2}{c}{Mean across FPRs}\\
        
        \cmidrule[\heavyrulewidth]{2-11}
        
        & Power
        & \multicolumn{1}{c|}{Reward}
        & Power
        & \multicolumn{1}{c|}{Reward}
        & Power
        & \multicolumn{1}{c|}{Reward}
        & Power
        & \multicolumn{1}{c|}{Reward}
        & Power                 
        & Reward\\
        
        \midrule
        
        \multicolumn{1}{c|}{TS-PostDiff}
        & 0.999
        & \multicolumn{1}{c|}{0.571}
        & 0.997
        & \multicolumn{1}{c|}{0.578} 
        & 0.995
        & \multicolumn{1}{c|}{0.582} 
        & 0.990
        & \multicolumn{1}{c|}{0.585} 
        & 0.995
        & 0.579 \\
        
        \cmidrule[\heavyrulewidth]{2-9}
        
        \multicolumn{1}{c|}{}
        & \multicolumn{2}{c|}{$\delta$ = 0.15}
        & \multicolumn{2}{c|}{$\delta$ = 0.125}
        & \multicolumn{2}{c|}{$\delta$ = 0.1}
        & \multicolumn{2}{c|}{$\delta$ = 0.075}
        &
        &\\ 
        
        \midrule
        
        \multicolumn{1}{c|}{}
        & 0.998
        & \multicolumn{1}{c|}{0.563} 
        & 0.993
        & \multicolumn{1}{c|}{0.572} 
        & 0.980
        & \multicolumn{1}{c|}{0.580} 
        & 0.936
        & \multicolumn{1}{c|}{0.586} 
        & 0.977
        & 0.575\\
        
        \cmidrule[\heavyrulewidth]{2-9}
        
         \multicolumn{1}{c|}{$\epsilon$/TT-TS}
        & \multicolumn{2}{c|}{$\beta$ = 0.8375}
        & \multicolumn{2}{c|}{$\beta$ = 0.8875}
        & \multicolumn{2}{c|}{$\beta$ = 0.9375}
        & \multicolumn{2}{c|}{$\beta$ = 0.975}
        &
        &\\ 
        
        \multicolumn{1}{c|}{}
        & \multicolumn{2}{c|}{$\epsilon$ = 0.325}
        & \multicolumn{2}{c|}{$\epsilon$ = 0.225}
        & \multicolumn{2}{c|}{$\epsilon$ = 0.125}
        & \multicolumn{2}{c|}{$\epsilon$ = 0.05}
        &
        &\\
        
        \midrule
        
        \multicolumn{1}{c|}{TS-ProbClip}
        & 0.992
        & \multicolumn{1}{c|}{0.564} 
        & 0.974
        & \multicolumn{1}{c|}{0.576} 
        & 0.938
        & \multicolumn{1}{c|}{0.582} 
        & 0.895
        & \multicolumn{1}{c|}{0.586} 
        & 0.950
        & 0.577 \\
        
        \cmidrule[\heavyrulewidth]{2-9}
        
        \multicolumn{1}{c|}{}
        & \multicolumn{2}{c|}{prob\_max = 0.8375}
        & \multicolumn{2}{c|}{prob\_max = 0.9}
        & \multicolumn{2}{c|}{prob\_max = 0.9375}
        & \multicolumn{2}{c|}{prob\_max = 0.9625}
        &
        &\\
        
        \bottomrule

        \toprule

        \multicolumn{11}{c}{arm difference = 0.3}\\
        
        \midrule
        
        \multicolumn{1}{c|}{Algorithm}
        & \multicolumn{5}{c|}{Power}
        & \multicolumn{5}{c}{Reward}\\
        
        \midrule
        
        \multicolumn{1}{c|}{Uniform Random}
        & \multicolumn{5}{c|}{1.000}
        & \multicolumn{5}{c}{0.500}\\
        
        \midrule
        
        \multicolumn{1}{c|}{TS}
        & \multicolumn{5}{c|}{0.980}
        & \multicolumn{5}{c}{0.642}\\
        
        \midrule
        
        \multicolumn{1}{c|}{Algorithm}
        & \multicolumn{2}{c|}{FPR = 0.06}
        & \multicolumn{2}{c|}{FPR = 0.07}
        & \multicolumn{2}{c|}{FPR = 0.08}
        & \multicolumn{2}{c|}{FPR = 0.1}
        & \multicolumn{2}{c}{Mean across FPRs}\\
        
        \cmidrule[\heavyrulewidth]{2-11}
        
        & Power
        & \multicolumn{1}{c|}{Reward}
        & Power
        & \multicolumn{1}{c|}{Reward}
        & Power
        & \multicolumn{1}{c|}{Reward}
        & Power
        & \multicolumn{1}{c|}{Reward}
        & Power                 
        & Reward\\
        
        \midrule
        
        \multicolumn{1}{c|}{TS-PostDiff}
        & 1.000
        & \multicolumn{1}{c|}{0.634}
        & 1.000
        & \multicolumn{1}{c|}{0.636} 
        & 0.999
        & \multicolumn{1}{c|}{0.638} 
        & 0.998
        & \multicolumn{1}{c|}{0.640} 
        & 0.999
        & 0.637 \\
        
        \cmidrule[\heavyrulewidth]{2-9}
        
        \multicolumn{1}{c|}{}
        & \multicolumn{2}{c|}{$\delta$ = 0.15}
        & \multicolumn{2}{c|}{$\delta$ = 0.125}
        & \multicolumn{2}{c|}{$\delta$ = 0.1}
        & \multicolumn{2}{c|}{$\delta$ = 0.075}
        &
        &\\ 
        
        \midrule
        
        \multicolumn{1}{c|}{}
        & 1.000
        & \multicolumn{1}{c|}{0.598} 
        & 1.000
        & \multicolumn{1}{c|}{0.612} 
        & 1.000
        & \multicolumn{1}{c|}{0.626} 
        & 0.994
        & \multicolumn{1}{c|}{0.636} 
        & 0.998
        & 0.618\\
        
        \cmidrule[\heavyrulewidth]{2-9}
        
         \multicolumn{1}{c|}{$\epsilon$/TT-TS}
        & \multicolumn{2}{c|}{$\beta$ = 0.8375}
        & \multicolumn{2}{c|}{$\beta$ = 0.8875}
        & \multicolumn{2}{c|}{$\beta$ = 0.9375}
        & \multicolumn{2}{c|}{$\beta$ = 0.975}
        &
        &\\ 
        
        \multicolumn{1}{c|}{}
        & \multicolumn{2}{c|}{$\epsilon$ = 0.325}
        & \multicolumn{2}{c|}{$\epsilon$ = 0.225}
        & \multicolumn{2}{c|}{$\epsilon$ = 0.125}
        & \multicolumn{2}{c|}{$\epsilon$ = 0.05}
        &
        &\\
        
        \midrule
        
        \multicolumn{1}{c|}{TS-ProbClip}
        & 1.000
        & \multicolumn{1}{c|}{0.599} 
        & 1.000
        & \multicolumn{1}{c|}{0.617} 
        & 0.997
        & \multicolumn{1}{c|}{0.628} 
        & 0.985
        & \multicolumn{1}{c|}{0.634} 
        & 0.996
        & 0.620 \\
        
        \cmidrule[\heavyrulewidth]{2-9}
        
        \multicolumn{1}{c|}{}
        & \multicolumn{2}{c|}{prob\_max = 0.8375}
        & \multicolumn{2}{c|}{prob\_max = 0.9}
        & \multicolumn{2}{c|}{prob\_max = 0.9375}
        & \multicolumn{2}{c|}{prob\_max = 0.9625}
        &
        &\\
        
        \bottomrule

    \end{tabular}
    \caption{Comparison of Power and Mean Reward for Uniform Random, TS, TS-PostDiff, $\epsilon$/TT-TS, and TS-ProbClip with sample size = 785 and three arm differences (0.1, 0.2, and 0.3). For TS-PostDiff, $\epsilon$/TT-TS, and TS-ProbClip, we choose four different parameters for each of them that gives a false positive rate of 0.06, 0.07, 0.08 and 0.1, respectively. 
    }
    \label{tab:name2}
\end{table*}

\end{appendices}